\documentclass[11pt,a4paper]{article}
\usepackage[hyperref]{rocling2025}
\usepackage{times}
\usepackage{latexsym}

\usepackage{microtype}

\usepackage{xcolor}         
\usepackage{amsmath}
\usepackage{amsthm}
\usepackage{amssymb}
\usepackage{mathtools}
\usepackage{xparse}
\usepackage{graphicx}
\usepackage{caption}
\usepackage{array}
\usepackage{multirow}
\usepackage{arydshln}
\usepackage{makecell}
\usepackage{booktabs}
\usepackage{diagbox}

\roclingfinalcopy 
\def\roclingpaperid{9} 

\title{LAUD: Integrating Large Language Models with Active Learning for Unlabeled Data}

\author{%
  Tzu-Hsuan Chou \\
  CMoney Technology Corporation \\
  \texttt{kevin\_chou@cmoney.com.tw} \\
  \And
  Chun-Nan Chou \\
  CMoney Technology Corporation \\
  \texttt{jason\_chou@cmoney.com.tw}
}

\date{}

\begin{document}
\maketitle

\begin{abstract}
Large language models (LLMs) have shown a remarkable ability to generalize beyond their pre-training data, and fine-tuning LLMs can elevate performance to human-level and beyond.
However, in \mbox{real-world} scenarios, lacking labeled data often prevents practitioners from obtaining \mbox{well-performing} models, thereby forcing practitioners to highly rely on \mbox{prompt-based} approaches that are often tedious, inefficient, and driven by trial and error.
To alleviate this issue of lacking labeled data, we present a learning framework integrating \textbf{L}LMs with \textbf{a}ctive learning for \textbf{u}nlabeled \textbf{d}ataset (LAUD).
LAUD mitigates the \mbox{cold-start} problem by constructing an initial label set with \mbox{zero-shot} learning.
Experimental results show that LLMs derived from LAUD outperform LLMs with \mbox{zero-shot} or \mbox{few-shot} learning on commodity name classification tasks, demonstrating the effectiveness of LAUD.
\end{abstract}


\section{Introduction}\label{sec:introduction}
Large language models (LLMs) have emerged as versatile tools which are capable of solving diverse tasks~\cite{2020t5,brown2020language,devlin-etal-2019-bert,radford2018improving,radford2019language}.
Through \mbox{zero-shot} learning~\cite{2020t5,brown2020language}, LLMs can perform downstream tasks without modifying parameters or architectures~\cite{radford2019language}.
As a form of transfer learning, \mbox{zero-shot} learning reformulates tasks to match LLM \mbox{pre-training} objectives, enabling knowledge transfer to new applications.
For example, \mbox{GPT-3}~\cite{brown2020language}, trained to predict the next token, can be adapted to binary classification by prompting \mbox{GPT-3} to output positive or negative.

Prior works~\cite{2020t5,brown2020language} show that the performance of \mbox{zero-shot} learning is left far behind \mbox{fine-tuning}.
But \mbox{fine-tuning} LLMs to achieve \mbox{human-level} performance often requires thousands or even millions of annotations~\cite{2020t5,brown2020language,devlin-etal-2019-bert,radford2018improving,radford2019language}.
Hence, reducing annotation cost remains a challenge and motivates continued research efforts.

There are two recent methods to tackle the issue of annotation cost: \mbox{few-shot} learning and \mbox{in-context} learning.
\mbox{Few-shot} learning leverages \mbox{pre-training} tasks of LLMs to tackle new tasks with only a few training examples~\cite{2020t5,brown2020language,schick-schutze-2021-exploiting,sun-etal-2022-nsp}.
On the other hand, \mbox{in-context} learning extends \mbox{zero-shot} learning by feeding demonstrations directly into the prompt~\cite{brown2020language}.
Empirical results~\cite{schick-schutze-2021-exploiting,sun-etal-2022-nsp} suggests that LLMs can adapt effectively under both approaches.

Traditionally, active learning~\cite{settles2009active} has been regarded as a promising approach in machine learning for reducing annotation costs.
Active learning strategically selects informative examples~\cite{LewisG94,xu2003representative} and queries oracles for annotations.
However, active learning always encounters the \mbox{so-called} \mbox{cold-start} situation~\cite{yuan-etal-2020-cold,jin2022cold}.
Prior works typically either assumed the existence of initial labeled sets~\cite{ein-dor-etal-2020-active} or randomly sample data points to assign annotations~\cite{LewisG94}.
While random sampling offers an intuitive solution to the \mbox{cold-start} issue, random sampling is prone to produce imbalanced distribution of the initial labeled sets~\cite{kothawade2021similar}
In practice, the \mbox{cold-start} issue forces practitioners to abandom active learning in favor of prompt engineering---an tedious, inefficient, and \mbox{trial-and-error} process that rarely yields robust performance~\cite{wenliang2025promptinghardunderstandingprompts,margatina2023activelearningprinciplesincontext}.
Consequently, the \mbox{cold-start} issue prevents practitioners from deriving LLMs for specific tasks via active learning~\cite{ein-dor-etal-2020-active,margatina2023activelearningprinciplesincontext}.
Throughout the remaining part of this paper, we refer to task-specific LLMs as TLLMs.

To alleviate the \mbox{cold-start} problem that hinders the derivation of TLLMs through active learning, we introduce a learning framework integrating \textbf{L}LMs with \textbf{a}ctive learning for \textbf{u}nlabeled \textbf{d}ata (LAUD).
Through active learning, LAUD iteratively acquires strategically selected annotations to guide \mbox{fine-tuning}, ultimately yielding a TLLM at minimal annotation cost.
In addition to reducing \mbox{fine-tuning} expenses, TLLMs produced by LAUD consistently outperform \mbox{zero-shot} and \mbox{few-shot} baselines, as demonstrated in our experiments.

In summary, this paper makes the following four key contributions:
\begin{enumerate}
    \item We propose LAUD: a learning framework integrating LLMs with active learning to transform unlabeled datasets into TLLMs.
    \item We address the \mbox{cold-start} problem by initializing active learning processes with \mbox{off-the-shelf} LLMs.
    \item Through experiments, we demonstrate that LAUD efficiently derives TLLMs that surpass \mbox{zero-shot} and \mbox{few-shot} baselines on commodity name classification.
    \item We validate the effectiveness of LAUD in a \mbox{real-world} \mbox{ad-targeting} system, where the TLLMs derived from LAUD yield substantial relative improvements in terms of \mbox{click-through} rate (CTR).
\end{enumerate}

\section{Methodology}\label{sec:method}

\begin{figure*}
  \centering
  \includegraphics[scale=0.6]{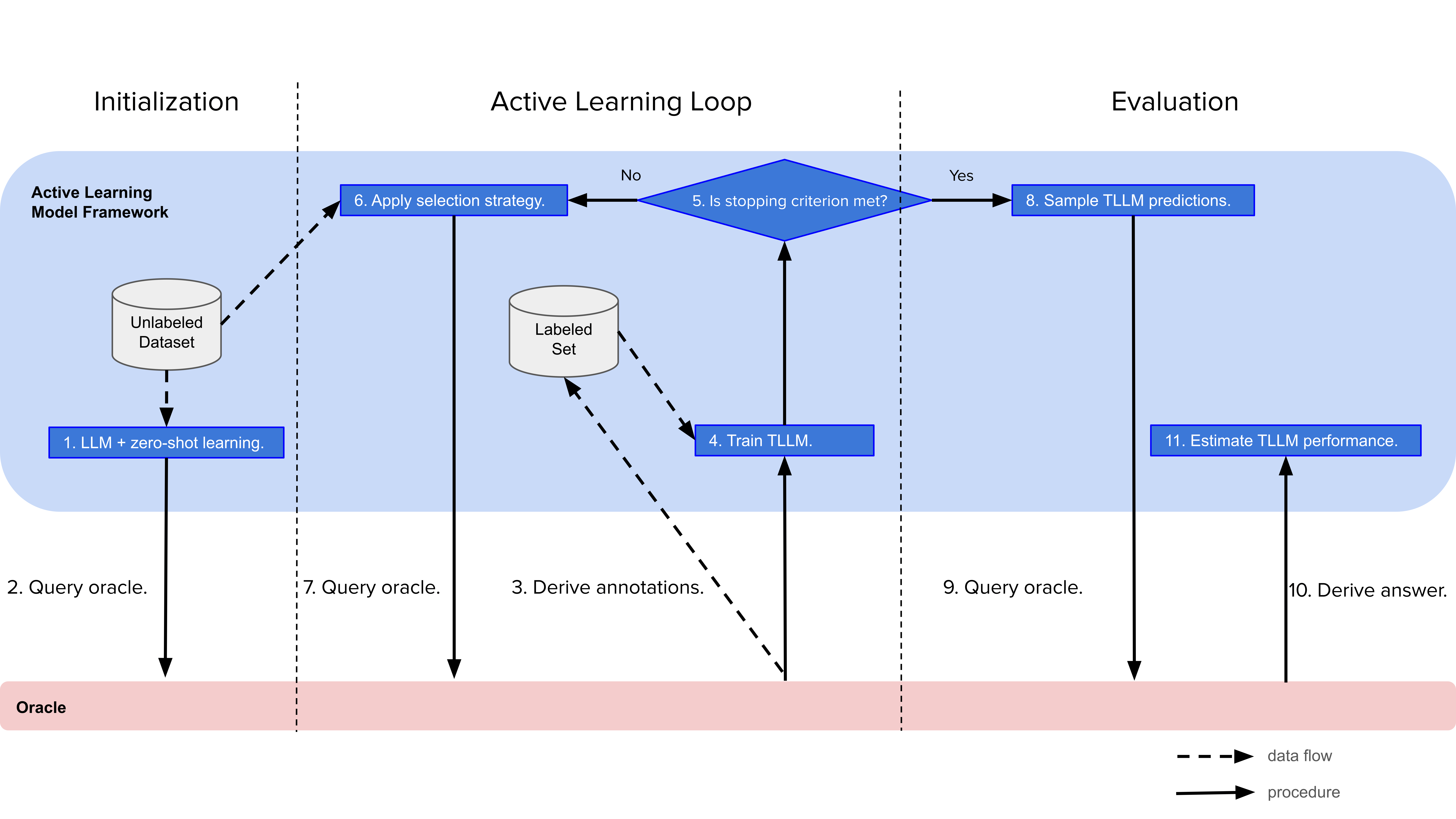}
  \caption{
    Illustration of LAUD.
    LAUD integrates LLMs with active learning to derive TLLMs from unlabeled data.
    One or more oracles in LAUD are queried to provide annotations for training and evaluating TLLMs.
  }
  \label{fig:method}
\end{figure*}

Because LAUD employs the active learning paradigm, we first elaborate on the active learning process in our learning framework, followed by the illustration of evaluation and oracle.
Figure~\ref{fig:method} depicts the overall skeleton of LAUD.

\subsection{Initialization}\label{sec:initialization}

In the initialization stage of LAUD, we use LLMs with \mbox{zero-shot} learning to tackle the \mbox{cold-start} issue of active learning since our input dataset is unlabeled.
First, we make the initial prediction for each data point via \mbox{zero-shot} learning.
Afterwards, based on the initial prediction, we collect annotations for each class.
Since \mbox{zero-shot} learning may lead to the potential of problematic predictions, we only consider the data points with high confidence.
By using our initialization procedure, we can derive an initial labeled set that is a balanced distribution without manually evaluating every data point.

\subsection{Active Learning Loop}\label{sec:collection}

In each iteration of this stage, LAUD firstly derives a TLLM by using the annotations that have been collected so far and include those annotations obtained during the initialization stage.
If the training cost is affordable, we can derive TLLMs through fine-tuning.
Otherwise, we can use \mbox{in-context} \mbox{few-shot} learning to transform a \mbox{task-agnostic} LLM into TLLM.

After a TLLM completes its training, the stopping criterion is evaluated.
If the stopping criterion is met, we stop the active learning loop.
Otherwise, we apply a selection strategy to find annotation candidates, and subsequently oracles provide the annotations for the selected candidates.
The same loop repeats to gather more annotations until the stopping criterion is met, and better TLLMs are desirable to emerge when annotations accumulate.

\subsection{Evaluation}\label{sec:evaluation}

After the active learning process is finished, we estimate the performance of the final TLLM.
Since as usual there are no test sets available in the active learning scenario, we sample the predictions made by the final TLLM and ask one or more oracles to provide the corresponding answers.
The final TLLM can be the TLLM obtained in the last active learning loop or can be other LLMs trained using all annotations derived from the active learning process.
In both cases, we exploit the final TLLM to gain the predictions of the unlabeled dataset and then request one or more oracles to examine the predictions.

\subsection{Oracle}\label{sec:oracle}

Depending on the requirement of annotation precision, an oracle that we refer in LAUD can be a person or an LLM.
When high annotation precision is desired, the oracle should be a person.
If the annotations with noise can be acceptable or tolerated, the oracle could be an LLM.

An oracle in LAUD plays three different roles as follows.
First, in the initialization stage, an oracle is queried to generate the initial labeled set based on the predictions of LLMs with \mbox{zero-shot} learning.
Second, in each iteration of the active learning loop, an oracle is queried to provide annotations for the selected data points that are used to train TLLMs.
Finally, an oracle helps estimate the performance of the final TLLM by examining the sampled predictions.

\section{Demonstration}\label{sec:demonstration}

To demonstrate the effectiveness of LAUD, this section presents both an experimental evaluation and a \mbox{real-world} application.
The three key findings are summarized as follows:
\begin{enumerate}
    \item TLLMs derived from LAUD outperform both TLLMs with random sampling and LLMs with \mbox{zero-shot} learning.
    \item Using an LLM as the oracle in LAUD can achieve the competitive performance with the human oracle.
    \item In a \mbox{real-world} \mbox{ad-targeting} system, deploying TLLMs derived from LAUD led to significant gains in CTR.
\end{enumerate}

\subsection{Experiments}\label{sec:experiments}

We use commodity name classification, i.e., commodity names are divided into different categories, as the demonstration experiments.

\subsubsection{Unlabeled Dataset}\label{sec:unlabeled-dataset}

Our unlabeled dataset consists of commodity names, which are mainly written in Chinese.
Initially, there are billions of commodity records in our unlabeled dataset.
To boost our evaluation, we filter out commodities that are purchased less frequently in our experiments, thereby remaining \(100\)k commodities (contribute \(> 60\%\) of transaction records).

Even though the essence of commodity name classification is a \mbox{multi-class} classification task, without loss of generality, we further reformulate commodity name classification as a binary classification task by asking LLMs whether a commodity belongs to a category.
Two categories ``coffee'' and ``tea'' are chosen as exemplars.

\subsubsection{Implementation}

This subsection illustrates an implementation of LAUD to demonstrate the capability of obtaining TLLMs that can solve the problem of commodity name classification.
We follow the narrative structure of Section~\ref{sec:method} to describe the details of our implementation.

\paragraph{Initialization}

We choose \mbox{NSP-BERT}~\cite{sun-etal-2022-nsp} as the LLM to perform \mbox{zero-shot} learning on the unlabeled dataset in all our experiments.
We regard commodity name classification task as a NSP (next sentence prediction) task~\cite{devlin-etal-2019-bert}.
For example, in order to predict whether a commodity name belongs to the coffee category, we can deal with the task by using the following prompt template
\begin{align*}
    & (S_1, S_2) \\
    & = (\text{``Commodity with name \{commodity\}''}, \\
    & \text{``is belong to coffee category.''}),
\end{align*}
and then ask \mbox{NSP-BERT} to predict whether \(S_2\) can be a next sentence of \(S_1\).
If the answer is positive, \mbox{NSP-BERT} should output \(1\).
Otherwise, \mbox{NSP-BERT} should output \(0\).
In all our experiments, we use the \texttt{bert-base-chinese}\footnote{\url{https://huggingface.co/google-bert/bert-base-chinese}} checkpoint.

In order to form the annotation candidates, we find \(K\) data points that \mbox{NSP-BERT} has the highest confidence in.
After the annotation candidates are formed, we ask an oracle to provide annotations.
Since commodity name classification is a binary classification task, we force \(K/2\) positive data points and \(K/2\) negative data points to guarantee a balanced distribution for easing the subsequent training.
In all our experiments, we set \(K\) to \(16\).
Note that \(K\) throughout this subsection refers to the same one. 

\paragraph{Active Learning Loop}

We choose \mbox{NSP-BERT}~\cite{sun-etal-2022-nsp} as the model architecture of TLLM in all our experiments.
In each iteration of the active learning loop, we use the same prompt template in the initialization stage of our experiments and \mbox{fine-tune} \mbox{NSP-BERT} with the annotations accumulated so far.
We use Adam~\cite{diederik2014adam} as the optimizer with learning rate of \(1e-5\), \(\beta_1 = 0.9\), \(\beta_2 = 0.999\), L2 weight decay of \(0.01\).
The \mbox{fine-tuning} procedure is implemented by using PyTorch~\cite{pytorch}.

We use \mbox{pool-based} uncertainty sampling~\cite{holub2008entropy} as the selection strategy.
In each iteration, we retrieve \(K\) data points that the \mbox{fine-tuned} \mbox{NSP-BERT} is most uncertain about and ask an oracle to provide annotations.

Regarding the stopping criterion, we stop the active learning loop when the maximum number of iterations is reached.
In all our experiments, we set the maximum number of iterations to \(9\), which causes \(160\) annotations in total.
The reason why the number of annotations is \(160\) instead of \(144\) is because we include the annotations derived in the initialization stage.

\paragraph{Evaluation}

We use precision to estimate TLLMs since we tackle the binary classification task.
We sample \(N\) data points inferred to be positive and ask an oracle to provide the corresponding ground truth.
\(N\) is set to \(200\) throughout all our experiments.

\begin{align*}
    & \operatorname{Precision} = \frac{\text{\#True-Positive in population}}{\text{\#Inferred-Positive in population}} \\
    & \quad \approx \frac{\text{\#True-Positive in sample}}{\text{\#Inferred-Positive in sample}} \\
    & \quad = \frac{\text{\#True-Positive in sample}}{N}.
\end{align*}

\paragraph{Oracle}

We compare human oracles to LLM oracles.
In each experiment, only one oracle is involved in the learning process.
In all our experiments, the human oracle is the same person, and the LLM oracle is Gemini.
More specifically, we use \texttt{gemini-flash-1.5-001}\footnote{\url{https://cloud.google.com/vertex-ai/generative-ai/docs/models/gemini/1-5-flash}} checkpoint for Gemini.

\subsubsection{Baseline}

To show the effectiveness of LAUD, we compare \mbox{NSP-BERT} derived from LAUD (denoted as \mbox{TLLM + LAUD}) to \mbox{NSP-BERT} with \mbox{zero-shot} learning (denoted as \mbox{LLM + ZL}).
In order to have a fair comparison, we use the same prompt template for all settings.

To evaluate the impact of active learning, we also compare \mbox{TLLM + LAUD} and \mbox{NSP-BERT} with randomly selected positive samples (denoted as \mbox{TLLM + RAND}).
To derive annotations \mbox{without} active learning, we use the results of \mbox{zero-shot} learning to derive annotation candidates that \mbox{NSP-BERT} is highly confident in and ask a person to provide annotations.

\subsubsection{Results}

\begin{table*}
\centering
\begin{tabular}{|c|c|c|c|}
\hline
\multicolumn{4}{|c|}{Category: coffee} \\
\hline
Method      & Oracle & Estimated Precision & \#Inferred-Positive \\
\hline
LLM + ZL    & Human  & 18.0\%              & 60,649              \\
\hline
TLLM + RAND & Human  & 96.5\%              & 1,102               \\
\hline
TLLM + LAUD & Human  & \textbf{98.0\%}     & 4,249               \\
\hline
TLLM + LAUD & LLM    & 95.0\%              & 2,535               \\
\hline
\end{tabular}
\caption{
    Estimated performance of \mbox{coffee-categorization} task.
    \#\mbox{Inferred-Positive} means the number of positive data points inferred by TLLM.
}
\label{tab:coffee}
\end{table*}

\begin{table*}
\centering
\begin{tabular}{|c|c|c|c|}
\hline
\multicolumn{4}{|c|}{Category: tea} \\
\hline
Method      & Oracle & Estimated Precision & \#Inferred-Positive \\
\hline
LLM + ZL    & Human  & 20.5\%              & 46,111              \\
\hline
TLLM + RAND & Human  & 83.0\%              & 4,253               \\
\hline
TLLM + LAUD & Human  & \textbf{95.0\%}     & 4,320               \\
\hline
TLLM + LAUD & LLM    & 93.0\%              & 8,434               \\
\hline
\end{tabular}
\caption{
    Estimated performance of \mbox{tea-categorization} task.
    \#\mbox{Inferred-Positive} means the number of positive data points inferred by TLLM.
}
\label{tab:tea}
\end{table*}

Our experiments yield three key insights:

\begin{enumerate}
    \item The effectiveness of TLLMs: TLLMs consistently outperform \mbox{zero-shot} LLMs, including strong commercial APIs such as \mbox{GPT-4o-mini}.
    \item The impact of active learning: Combining \mbox{fine-tuning} with informative data selection improves precision by up to 12\%.
    \item LLMs as oracles are possible: \mbox{LLM-based} oracles achieve precision comparable to human oracles, offering a \mbox{cost-efficient} alternative for annotation.
\end{enumerate}

\paragraph{The effectiveness of TLLM}

In both the coffee and tea experiments (see rows three to five in both Tables~\ref{tab:coffee} and \ref{tab:tea}), we found that TLLMs, even without active learning, have significantly outperformed LLMs with \mbox{zero-shot} learning (the third row).
The huge difference suggests that while \mbox{off-the-shelf} LLMs can perform adequately across a range of tasks, their performance may not be competitive to models \mbox{fine-tuned} for specific tasks.

In additional experiments, we employed \mbox{GPT-4o-mini}\footnote{\url{https://platform.openai.com/docs/models/gpt-4o-mini}} with \mbox{zero-shot} learning, denoted as \mbox{GPT + ZL}, for commodity name classification.
\mbox{GPT + ZL} achieved a precision of approximately 60\%, representing a notable improvement over \mbox{NSP-BERT} (the third row).
However, TLLMs attaining at least 80\% precision consistently outperformed \mbox{GPT + ZL}.
These findings indicate that while commercial LLM APIs offer strong and easily accessible baselines for various tasks, their performance remains inferior to TLLMs.

\paragraph{The impact of active learning}

In both the coffee and tea experiments (see rows four and five in both Tables~\ref{tab:coffee} and \ref{tab:tea}), we found that \mbox{TLLM + LAUD} (the fifth row) has the best precision among the other methods.
The difference between TLLM with and without active learning can be up to \(12\%\).
The results suggest that the combination of \mbox{task-specific} \mbox{fine-tuning} and active learning creates a synergistic effect, where the model not only learns from the most relevant data but also from the most informative data.

\paragraph{LLMs as oracles are possible}\label{sec:zero-shot}

In both the coffee and tea experiments (see rows five and six in both Tables~\ref{tab:coffee} and \ref{tab:tea}), we found that using LLMs as oracles instead of human oracles does not make much difference in terms of precision.
The results suggest that LLMs may serve as one kind of effective surrogates for human oracles in the active learning process.
The comparable performance indicates that LLM oracles are capable of providing \mbox{high-quality} annotations during the learning process.
However, the two categories in our experiments show differences in the number of inferred positive data points.
The phenomenon is still not clear to us and remains to be explored.

\subsection{Applications}\label{sec:applications}

In addition to the experiments in Section~\ref{sec:experiments}, we futher used a \mbox{real-world} scenario to assess the effectiveness of TLLMs derived from LAUD.
As a case study, we considered an \mbox{ad-targeting} system built on commodity classification.
The original algorithm in the ad-targeting system relies on keyword-based commodity classifiers.
We replaced this baseline with TLLMs derived from LAUD and compared their performance in a controlled A/B testing setup.
The test results showed that predictions generated by \mbox{TLLM + LAUD} achieved approximately a 50\% relative improvement over the \mbox{keyword-based} classifiers in terms of CTR.
The promising results indicate that LAUD extends beyond experimental evaluation, proving to be applicable and impactful in \mbox{real-world} contexts.

\section{Conclusions}

Our proposed LAUD integrates LLMs with active learning to derive TLLMs for unlabeled data.
As illustrated in the exemplary commodity name classification, LAUD mitigates the \mbox{cold-start} problem with \mbox{zero-shot} learning and yields TLLMs that significantly outperform LLMs without \mbox{task-specific} knowledge.
Additionally, LLMs show the potential to be one of the alternatives to human oracles and thus open the possibility of automating and scaling the annotation process.
When deployed in a \mbox{real-world} \mbox{ad-targeting} system, TLLMs generated by LAUD achieved significant improvements in CTR.
Nevertheless, as future works, variations in inferred positive data points across different commodity categories are required for further investigations into the interplay between TLLMs, active learning, and oracle selection.
Regarding commodity name classification, there still remains one question: whether a category we consider is scarce or not causes any impact on our proposed learning framework.



\bibliography{rocling2025}
\bibliographystyle{acl_natbib}

\end{document}